\documentclass[aps,prx,twocolumn,superscriptaddress,a4paper,english,longbibliography]{revtex4-1}

\usepackage{times}
\usepackage{color}
\usepackage[colorlinks=true,urlcolor=blue,citecolor=blue]{hyperref}
\usepackage{url}
\usepackage{breakurl}
\usepackage{soul}
\usepackage{graphicx}
\usepackage{amsmath}
\usepackage{subfigure}
\usepackage{xcolor}
\usepackage{amssymb}
\usepackage{latexsym}
\usepackage{hyperref}
\DeclareGraphicsExtensions{.png}
\usepackage{float}

\usepackage{xcolor}
\usepackage[urlcolor=blue]{hyperref}      
\hypersetup{
    colorlinks = true,                    
    citecolor = {blue},
    linkcolor = {purple},
           }
\begin{document}	
	
\title{Bayesian Tensor Network with Polynomial Complexity for Probabilistic Machine Learning} 

\author{Shi-Ju Ran}\email[Email: ]{sjran@cnu.edu.cn}
\affiliation{Department of Physics, Capital Normal University, Beijing 100048, China}
\date{\today}

\begin{abstract}
 	It is known that describing or calculating the conditional probabilities of multiple events is exponentially expensive. In this work, Bayesian tensor network (BTN) is proposed to efficiently capture the conditional probabilities of multiple sets of events with polynomial complexity. BTN is a directed acyclic graphical model that forms a subset of TN. To testify its validity for exponentially many events, BTN is implemented to the image recognition, where the classification is mapped to capturing the conditional probabilities in an exponentially large sample space. Competitive performance is achieved by the BTN with simple tree network structures. Analogous to the tensor network simulations of quantum systems, the validity of the simple-tree BTN implies an ``area law'' of fluctuations in image recognition problems.
\end{abstract}

\maketitle

\section{Introduction}

Tensor network (TN) is a powerful non-linear (more specifically, multi-linear) model \cite{RTPC+17TNrev} that was originally developed in the field of quantum many-body physics \cite{VMC08MPSPEPSRev, CV09TNSRev, O19TNrev}. With the striking resemblances to neural network (NN), TN has recently been applied to machine learning \cite{SS16TNML, LRWP+17MLTN, HWFWZ17MPSML, GJLP18MPOML, S18MERAML, CWXZ19generateTTNML}. Treated as quantum many-body states or operators \cite{GPC18gTNML, GSPEC19PTNML}, TN is expected to provide a natural solution for probabilistic machine learning \cite{G15PMLrev}. Therefore, TN is also called Born machine \cite{cheng2018information}, as the quantum counterpart of the famous Boltzmann machine.

Aside from the probabilistic interpretation, one significant advantage of TN is its efficiency. For quantum many-body simulations, the complexity of a many-body state and the dimension of the Hilbert space scale exponentially with the system size. TN ``magically'' reduces the complexity to only scale polynomially \cite{O14TNSRev}. The validity of TN is based on the area law of the entanglement entropy, where the important correlations or fluctuations are short-range \cite{BZ17entbook, ECP10AreaLawRev}. Then TN uses only polynomially expensive resources to accurately capture the short-range fluctuations and obtains high accuracies.

This work concerns the conditional probabilities of multiple event, for which we have another well-established probabilistic model known as Bayesian belief network (BBN) \cite{FGG97BayesNet}. Classifications and inferences can be done with BBN by implementing Gibbs sampling after optimizing the network structure from the data. In this way, the causal relations can be inferred from the conditional probabilities among the nodes. However, either calculating the conditional probabilities of multiple events or obtaining the global optimal structure of the BBN is exponentially expensive or NP-hard \cite{DL94BayesNP, D96BayesNP}. One solution is to combine BBN with the ideas of deep learning, which has a long history \cite{barber1998ensemble, neal2012bayesian} and is currently under hot debate (see, e.g., \cite{OSKJ+19Bayes, SV19DeepInfer}).

\begin{figure}[tbp]
	\includegraphics[width=1\linewidth]{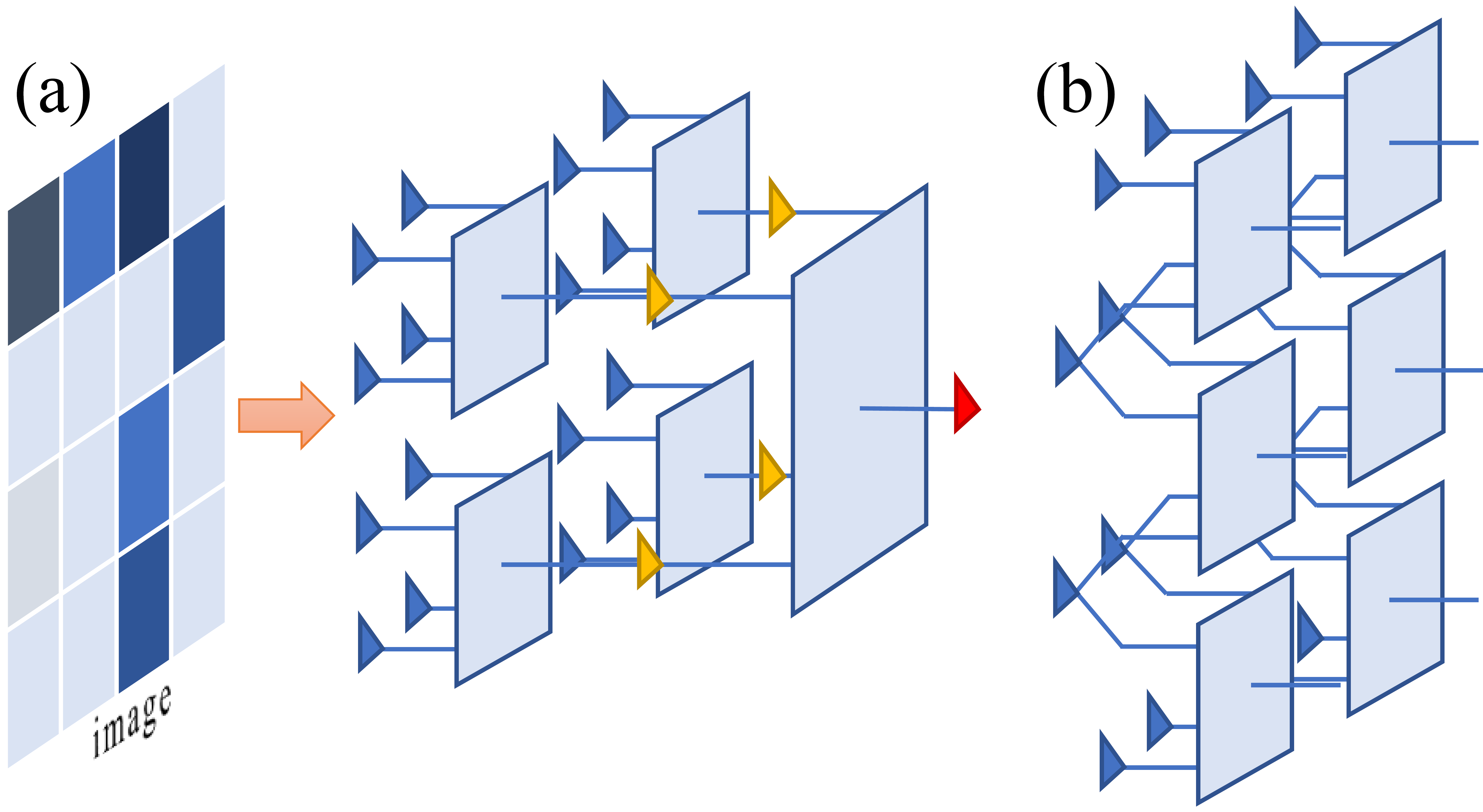}
	\caption{\label{Fig-BTN}  (Color online) (a) The illustration of image recognition by a simple-tree BTN of two layers. A ($4 \times 4$) image is mapped to ($4 \times 4$) sets of events as the root sets. The first layer contains four tensors that map the root sets (blue) to ($2 \times 2$) hidden sets (yellow).  The last layer contains one tensor that maps the hidden sets to the leaf set (red) that indicates the probability distribution of the classification of the image. (b) By using the tail-to-tail structure, the first layer can be designed to contain ($2 \times 3$) tensors, which map the root sets to ($2 \times 3$) hidden sets.}
\end{figure}

In this work, Bayesian tensor network (BTN) is proposed for probabilistic machine learning. BTN is a directed acyclic graphical model that forms a subset of TN (see a simple-tree BTN in Fig. \ref{Fig-BTN} (a) as an example); it inherits the advantages of both BBN and TN. BTN provides an efficient representation for the conditional probabilities of massive events with only polynomial complexity [see, for instance, Eqs. (\ref{eq-Complx}) and (\ref{eq-Complx2})]. The structure of a BTN is priorly designed, and the performance does not severely depend on the structure. The casual relations among the events can be inferred by calculating the conditional probabilities using tensor contractions instead of Gibbs samplings. A rotation optimization method \cite{SRG19preperation} is suggested to update BTN, which avoids gradient vanishing problem and exhibits high efficiency. To testify BTN, the image recognition is mapped to a probabilistic problem, i.e., to capture the conditional probabilities in an exponentially large sample space. BTN exhibits competitive performances on the fashion-MNIST dataset \cite{fMNIST}. The expressive power of BTN is discussed from the perspectives of the probabilities of events, the dimension of the vector space where the BTN is defined, and the ``area law'' \cite{B73EntAreaLaw, S93Sarea, PEDC05EntAreaLaw, ECP10AreaLawRev, F13arealawTRG, O14TNSRev} of the fluctuations of TN.

\section{Bayesian tensor network: definitions and algebras}

Before define Bayesian TN, let me firstly define Bayesian tensor. Considering a set (denoted as $X$) of mutually exclusive events (denoted as $x_i$), a vector $V$ is used to describes the probabilities of these events as the following. $V$ is positively defined, and its $i$-th element gives the probability of the $i$-th event as $V_i = P(x_i)$. We have
\begin{eqnarray}
\label{eq-P0}
\sum_i P(x_i) = \sum_i V_i \overset{\text{def}}{=} |V|_1 = 1.
\end{eqnarray}

\begin{figure}[tbp]
	\includegraphics[width=0.95\linewidth]{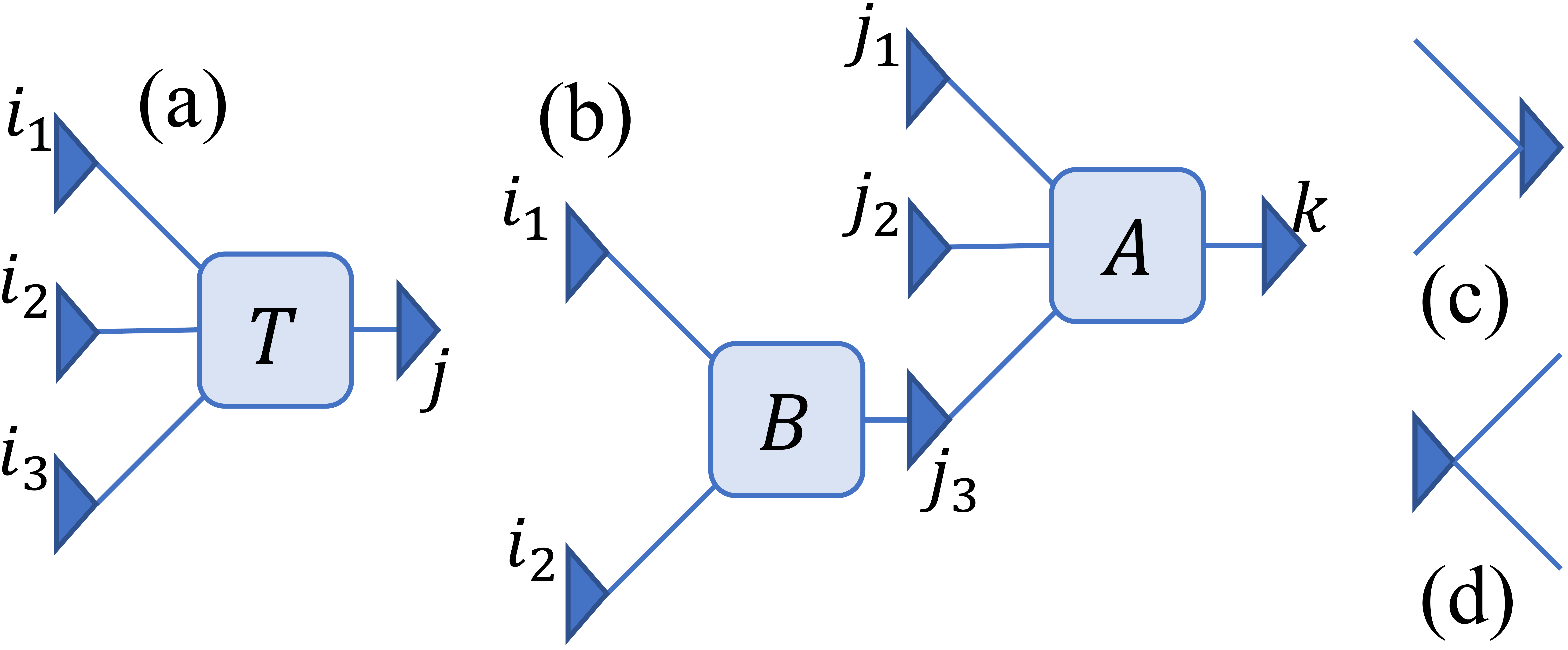}
	\caption{\label{Fig-BT} (Color online) Illustrations of (a) a Bayesian tensor $T_{j, i_1 i_2 i_3}$, (b) the contraction of two Bayesian tensors $\sum_{j_3} A_{k, j_1j_2j_3} B_{j_3, i_1 i_2}$, (c) head-to-head structure, and (d) tail-to-tail structure.}
\end{figure}

A \textit{Bayesian tensor} $T$ (see several related works by tensor decompositions at \cite{TSLS+08bayest, STS09bayest, XCH+10bayest, ZZC15bayest}) describes the conditional probabilities of multiple sets of mutually exclusive events. Here, let us consider ($\tilde{m}+1$) sets of events $x^{[m]}_{i_m} \in X^{[m]}$ ($m=1, 2, \cdots, \tilde{m}$) and $y_j \in Y$. We have
\begin{eqnarray}
\label{eq-condP}
P(y_j|x^{[1]}_{i_1}, x^{[2]}_{i_2}, \cdots, x^{[\tilde{m}]}_{i_{\tilde{m}}}) = T_{j, i_1 i_2 \cdots i_{\tilde{m}}},
\end{eqnarray}
where $T$ is a ($\tilde{m}+1$)-th ordered tensor and $j, i_1, i_2, \cdots, i_{\tilde{m}}$ are its indexes. The dimension of each index equals to the number of events in the corresponding set, i.e., $\dim(i_m) = \#(X^{[m]})$. A Bayesian tensor has a direction from $\{x^{[m]}_{i_m}\}$ to $y_j$. In Fig. \ref{Fig-BT}, we use squares to represent a Bayesian tensor, and use the bonds connecting with a square to represent the indexes. On each bond, we put a solid triangle to represent the set of events and to indicate the direction. The indexes $i_1, i_2, \cdots, i_{\tilde{m}}$ are dubbed as \textit{in-going indexes}; $j$ is dubbed as \textit{out-going index}.

To describe the conditional probabilities, $T$ is required to satisfy the following properties: (a) positivity: $T$ is positively defined. This is a direct result from Eq. (\ref{eq-condP}); (b) normalization: for any in-going indexes ($i_1, i_2, \cdots, i_{\tilde{m}}$), it has
\begin{eqnarray}
\label{eq-Tcond1}
\sum_j T_{j, i_1 i_2 \cdots i_{\tilde{m}}} \overset{\text{def}}{=} |T_{:, i_1 i_2 \cdots i_{\tilde{m}}}|_1 = 1,
\end{eqnarray}
where ``$:$'' means to go through all values of the corresponding index. Eq. (\ref{eq-Tcond1}) means $\sum_{y_j} P(y_j|x^{[1]}_{i_1}, x^{[2]}_{i_2}, \cdots, x^{[\tilde{m}]}_{i_{\tilde{m}}}) = 1$.

A tensor satisfying (b) must have the following property: (b$'$) for any vectors $V^{[m]}$ with $|V^{[m]}|_1=1$ and $m = 1, \cdots, \tilde{m}$, we have $|\mathcal{V}|_1 = 1$ with
\begin{eqnarray}
\label{eq-Tcond2}
\mathcal{V}_j = \sum_{ i_1 i_2 \cdots i_{\tilde{m}}}  T_{j, i_1 i_2 \cdots i_{\tilde{m}}} V^{[1]}_{i_1} V^{[2]}_{i_2} \cdots V^{[\tilde{m}]}_{i_{\tilde{m}}}.
\end{eqnarray}
When $T$ and all $\{V^{[m]}\}$ are positively defined, Eq (\ref{eq-Tcond2}) is equivalent to the total probability theorem $P(y_j) = \sum_{\{x^{[m]}_{i_m}\}} P(y_j|x^{[1]}_{i_1}, x^{[2]}_{i_2}, \cdots, x^{[\tilde{m}]}_{i_{\tilde{m}}}) \prod_m P(x^{[m]}_{i_m})$ with $P(x^{[m]}_{i_m}) = V^{[m]}_{i_m}$. Then $|\mathcal{V}|_1 = 1$ means $\sum_{j} P(y_j) =1$. Note it is assumed here that each two events from different sets (corresponding to the in-going indexes) are independent, i.e., $P(x^{[m]}_i, x^{[n]}_j) = P(x^{[m]}_i) P(x^{[n]}_j)$ when $x^{[m]}_i \in X^{[m]}$ and $x^{[n]}_j \in X^{[n]}$ with $m \neq n$. If two sets contain non-independent events, one needs to replace the products by the joint probability distributions.


(c) Direction inversion. Knowing the probability distribution of the sets, the directions of the indexes of a Bayesian tensor can be inverted by using Bayes' equation. For instance, consider three sets of events ($\{x^{[1]}\}$, $\{x^{[2]}\}$, and $\{x^{[3]}\}$) and the Bayesian tensor giving the conditional probabilities as $P(x^{[1]}_i| x^{[2]}_j, x^{[3]}_k) = T_{i,jk}$. Assume two probability distributions $P(x^{[2]}_j) = V^{{[2]}}_j$ and $P(x^{[3]}_k) = V^{[3]}_k$ are priorly known ($\{x^{[2]}\}$ and $\{x^{[3]}\}$ are independent to each other). We have the total joint probability distribution $P(x^{[1]}_i, x^{[2]}_j, x^{[3]}_k) = \tilde{T}_{ijk} = T_{i,jk} V^{[2]}_j V^{[3]}_k$. Then we have the probability distribution of $\{x^{[1]}\}$ as $P(x^{[1]}_i) = V^{[1]}_i = \sum_{jk} \tilde{T}_{ijk}$ [also see Eq. (\ref{eq-Tcond2})], and the joint distribution $P(x^{[1]}_i, x^{[2]}_j) = \sum_{k} \tilde{T}_{ijk}$. From Bayes' equation, the direction inversion is done as
\begin{eqnarray}
P(x^{[2]}_j| x^{[1]}_i, x^{[3]}_k) = T_{j, ik} = \frac{T_{i,jk}  V^{[2]}_j}{\sum_{j} T_{i,jk} V^{[2]}_j}, \label{eq-Inverse1} \\
P(x^{[3]}_k| x^{[1]}_i, x^{[2]}_j) = T_{k,ij} = \frac{T_{i,jk}  V^{[3]}_k}{\sum_{k} T_{i,jk} V^{[3]}_k} \label{eq-Inverse2}.
\end{eqnarray}
The equations above are the tensor versions of Bayes' equation.

The definition of \textit{Bayesian tensor network} (BTN) is very simple: the contractions of multiple Bayesian tensors. BTN is a well-defined subset of TN. This is different from the generalized TN's such as string-bond states which contain non-multi-linear operations \cite{SWVC08QMCTN, GPC18gTNML, GPARC18MLstringbond}. This is the reason why BTN does not require samplings. 

BTN can be deep. Graphically, we use a shared bond with a triangle on it to represent the same event in different Bayesian tensors. To proceed, let me introduce the concepts of \textit{root sets}, \textit{leaf sets}, and \textit{hidden sets}. Following the directions of the triangles, the root and leaf sets are the events at the starting and ending points of the whole BTN, respectively. The rest are the hidden sets. For simplicity, we assume that the network graph of a BTN is connected. In the following, let me show several important properties of BTN.

(a) \textit{Connection rules}. Consider that a set appears at more than one Bayesian tensors. If this set corresponds to an in-going index for all the connected tensors, one should use the tail-to-tail structure to connect to the tensors; if this set corresponds to an out-going index for all the connected tensors, one should use the head-to-head structure to connect.

(b) \textit{Contraction rule}. If a set appears at two Bayesian tensors as in-going and out-going index, respectively, one should connect these two tensors by a bond; this index represents a hidden set and should be contracted in the TN computations. Note a hidden index is analogous to a virtual or ancillary bond in the TN terminology \cite{VMC08MPSPEPSRev}.

(c) \textit{Directed acyclicality}. BTN is a directed acyclic graphical model. This property is the same as the BBN. A \textit{circle} in a BTN or BBN is forbidden, meaning if one goes along the directions of the triangles, one cannot go back to the tensors that have been got through. If a circle appears in the BTN, the causal relations will become inconsistent. Note that \textit{loops} are defined as the loops of the network graph, same to the definition in the TN's. There are no constraints concerning the loops as long as the model is directed acyclic. Different structure with loops can be designed applying the connection and contraction rules.

(d) \textit{Contraction consistency}. Given a BTN formed by two Bayesian tensors $A_{k, j_1j_2 \cdots}$ and $B_{j_m, i_1i_2 \cdots}$, the tensor by contracting the shared index as $C_{k, j_1 \cdots j_{m-1} j_{m+1} \cdots i_1 i_2 \cdots} = \sum_{j_m} A_{k, j_1j_2 \cdots} B_{j_m, i_1i_2 \cdots}$ is a Bayesian tensor. Fig. \ref{Fig-BT} (b) illustrates an example. This can be readily proven using the probabilistic language of the Bayesian tensors. An inference can be made from the contraction consistency: (d$'$) A BTN can be written as one Bayesian tensor. This inference can be proven by contracting all the shared indexes (hidden sets) in the BTN. An equivalent description of (d$'$) is that a BTN represents the conditional probabilities between the leaf and root sets. 

(e) \textit{Relation between loop and independence}. Assume that different root sets are independent. If different in-going sets of one Bayesian tensor in the BTN are dependent, there must exist at least one loop between this tensor and the root sets. An equivalent expression is that if the BTN is loop-free, any different in-going sets in one tensor are independent.

(f) \textit{Probability distributions of hidden and leaf sets}. Assume the probability distributions of the root sets are known, and there exist no loops in the BTN. Given a BTN, the probability distributions of all hidden and leaf set can be calculated. This can be done by contracting the BTN from the roots to the leafs following the directions of the triangles. If loops appear, one will get joint probability distributions of multiple sets, instead of the probability distribution of each single set.

(g) \textit{Inferences}. When property (f) holds, the conditional probabilities among any two root sets can be calculated. This can be done by using Bayes' equation (see Eqs. (\ref{eq-Inverse1}) and (\ref{eq-Inverse2}) as examples) to invert the Bayesian tensors on the path connecting the two root sets. Note that for BBN, the network structure is optimized according to the data. The network structure of BTN is priorly determined. We only optimize the parameters in the Bayesian tensors, not the structure. For inferences, the conditional probabilities among the root sets can be calculated after the Bayesian tensors are optimized. Meanwhile, samplings are not needed for BTN. Instead, one should implement tensor contractions.

(h) \textit{Parameter complexity}. The parameter complexity of a BTN is defined as the total number of parameters of the tensors. The parameter complexity scales polynomially with the total number of sets and (approximately) the average number of events in one sets (i.e., the dimensions of the indexes). For different network structures, the parameter complexity will be slightly different. In comparison, the parameter complexity of the full conditional probabilities scales exponentially with the number of sets.

(i) \textit{Contraction complexity}. The contraction complexity of a BTN is to characterize the complexity brought by the loops of the network graph. Without performing any approximations (such as those in the TN contraction algorithms; see, e.g., \cite{LN07TRG, OV09CTMRG}), the complexity increases exponentially with the number of loops. This property is the same as the standard TN contractions. In addition, the loops also increase the cost to calculate and save the chains of gradients in the back propagation (if implemented).


\section{Probabilistic image recognition and rotation optimization algorithm}

To testify BTN, let me use it for supervised learning by taking image recognition as an example. The central task is to capture the conditional probabilities between the images and their classifications. The dimension of the sample space scales exponentially with the number of features.

The first step is to map each feature (pixel; let me denote the $m$-th pixel in the $n$-th image as $x^{[m, n]}$) to a set of exclusive events $\{x^{[m, n]}_i\}$. The probability distribution of these events is determined by the value of the pixel. One simplest example is the binary images, where the events are taking the pixel $x^{[m, n]}$ to be 0 (black) or 1 (white). Obviously a pixel cannot be black and white at the same time, thus these events should be mutually exclusive. For a given pixel $x^{[m, n]}$, we can set the probability distribution as $P(x^{[m, n]}=0) = 1-x^{[m, n]}$ and $P(x^{[m, n]}=1) = x^{[m, n]}$. Thus the vector $V^{[m,n]}$ to put at the root is two-dimensional that satisfies
\begin{eqnarray}
\label{eq-Fmap1}
V^{[m,n]} = [1-x^{[m, n]}, x^{[m, n]}].
\end{eqnarray}
The map from features to the probability distributions of sets of events, e.g., Eq. (\ref{eq-Fmap1}), is dubbed as the \textit{probabilistic map}.

For the gray-scale images with $0 \leq x^{[m, n]} \leq 1$, one choice is to use the following map, which has been used in TN machine learning \cite{SS16TNML}, to define $d$-dimensional $\tilde{V}^{[m,n]}$'s as
\begin{eqnarray}
\tilde{V}^{[m,n]}_i=\sqrt{\binom{d-1}{i-1}}[\cos(\frac{\pi }{2} x^{[m, n]})]^{d-{i}} [\sin(\frac{\pi}{2}x^{[m, n]})]^{i-1}. \nonumber \\
\label{eq-Fmap2}
\end{eqnarray}
It meas that one can control the number (i.e., $d$) of the mutually exclusive events in a set. With Eq. (\ref{eq-Fmap2}), one has $|\tilde{V}^{[m,n]}|_2=1$. Therefore, we have $V^{[m,n]} = (\tilde{V}^{[m,n]})^2$, so that $V^{[m,n]}_i = P(x^{[m, n]}_i)$ and $|V^{[m,n]}|_1= 1$.

With the probability distributions of the root sets using the probabilistic map, the next step is to use BTN to capture the conditional probabilities between the root and leaf sets. One simplest example of a tree BTN that has been used in TN mahine learning \cite{LRWP+17MLTN} is illustrated in Fig. \ref{Fig-BTN} (a). For classification, there is only one leaf set, where the number of the events equal to the number of classes. For a tree, the network graph has no loops. If any hidden set of events are known (fixed), the root sets will be blocked into two independent parts. Note loops can be designed by adding tail-to-tail structures. Fig. \ref{Fig-BTN} (b) shows an example, where the ($4 \times 4$) root sets are mapped to ($3 \times 2$) hidden sets after the first layer. This, however, will inevitably increase the contraction complexity (see property (i) of BTN).

The loss function for classification can be chosen as the cross entropy, same as the NN's or TN's. For a NN, one usually implements softmax on the outputs to satisfy the normalization of the probability. For a BTN, this is not necessary since the normalization is naturally fulfilled (see property (d) or (d$'$) of BTN).

The tensors in the BTN can be updated by the gradient methods, i.e., $T \leftarrow T - \eta \frac{\partial f}{\partial T}$ with $f$ the loss function and $\eta$ the learning rate. However, our data imply that the methods to control the gradients of NN's (e.g., Adam \cite{KB15Adam}) works badly for BTN. This is possibly due to the normalization constraints given in Eq. (\ref{eq-Tcond1}).

In the following, a gradient optimization method is proposed by taking advantage of the normalization constraints, which is dubbed as \textit{rotation optimization}. The central idea is to rotate the normalized vectors, where the learning rate is controlled by  the rotation angel \cite{SRG19preperation}. To this end, I introduce the ancillary tensor
\begin{eqnarray}
Q_{j, i_1i_2\cdots} = \sqrt{T_{j, i_1i_2\cdots}}.
\label{eq-Q}
\end{eqnarray}
As $T$ is positively defined, $Q$ should be real. We have $|Q_{:, i_1i_2 \cdots}|_2 = 1$. The rotation optimization is to update $Q$ instead of $T$.

Firstly, one calculates the gradient $G_{j, i_1i_2\cdots}$ of one ancillary tensor (say $Q_{j, i_1i_2\cdots}$) by back propagation as $G =  \frac{\partial f}{\partial Q}$. Note $G$ is also a tensor (not a Bayesian tensor) with the same order as $Q$ and $T$. Then calculate the components that are orthogonal to the normalized vectors in $Q$ as
\begin{eqnarray}
G_{j,i_1i_2\cdots} \leftarrow G_{j, i_1i_2\cdots} - Q_{j, i_1i_2\cdots} \sum_{j} G_{j, i_1i_2\cdots} Q_{j, i_1i_2\cdots}. 
\label{eq-OrtG}
\end{eqnarray}
We have the orthogonality between $G$ and $Q$ as
\begin{eqnarray}
\sum_{j} G_{j,i_1i_2\cdots} Q_{j, i_1i_2\cdots} = 0.
\label{eq-OrtGQ}
\end{eqnarray}
It means that with fixed in-going indexes ($i_1, i_2, \cdots$), the vector $G_{:,i_1i_2\cdots}$ is orthogonal to $Q_{:, i_1i_2\cdots}$. Then normalize $G$ as
\begin{eqnarray}
G_{:,i_1i_2\cdots} \leftarrow \frac{G_{:,i_1i_2\cdots}}{|G_{:,i_1i_2\cdots}|_2}.
\label{eq-NormalizeG}
\end{eqnarray}
This normalization does not change the orthogonality between $G$ and $Q$. Then update $Q$ by $G$ as
\begin{eqnarray}
Q_{:, i_1i_2\cdots} \leftarrow Q_{:, i_1i_2\cdots} - G_{:, i_1i_2\cdots} \tan \theta,
\label{eq-updateQ}
\end{eqnarray}
where $\theta$ is the rotation angel according to the triangular relation between $G$ and $Q$. 
Finally, normalized $Q$ as 
\begin{eqnarray}
Q_{:, i_1i_2\cdots} \leftarrow \frac{Q_{:, i_1i_2\cdots}}{|Q_{:, i_1i_2\cdots}|_2},
\label{eq-normalizeQ}
\end{eqnarray}
so that $T$ satisfies Eq. (\ref{eq-Tcond1}). The potential gradient vanishing problem are avoided by the rotation optimization, since the changes of the normalized vectors are strictly controlled by the rotation angel. 

\section{Discussions and results: expressive power and area law}

I choose fashion-MNIST dataset to testify \cite{fMNIST}. These images are gray-scale with the definition $28 \times 28$, and belong to ten categories. We have the number of root sets $M=28 \times 28 = 784$ and the dimension of the leaf set $N_c=10$. The network structure is chosen as a tree similar to Fig. \ref{Fig-BTN} (a) (also see \cite{LRWP+17MLTN}). For simplicity, assume that the dimension of every root index (i.e., the number of events in every root set) is the same, denoted by $d$; the dimension of every hidden index equals to $\chi$. Then the dimension of the sample space equals to $d^M$. Obviously, it is extremely challenging for BBN to handle such a problem. For the BTN, each Bayesian tensor is chosen to have four in-going indexes and one out-going index. The parameter complexity of such a BTN scales as
\begin{eqnarray}
\#(\text{BTN}_1) = \frac{d^4\chi M}{4} + \chi^5 (\frac{M}{4}-1)  + \chi^4N_c.
\label{eq-Complx}
\end{eqnarray}
Another network structure is also tried, where each tensor has two in-going indexes and one out-going index (one may find a similar tree structure in \cite{CWXZ19generateTTNML}). The complexity satisfies
\begin{eqnarray}
\#(\text{BTN}_2) = \frac{d^2\chi M}{2} + \chi^3 (\frac{M}{2}-1)  + \chi^2N_c.
\label{eq-Complx2}
\end{eqnarray}
The complexities of both BTNs's scale polynomially with the number of sets of events ($M$) and the numbers of events in each set ($d$ and $\chi$). Note one can access larger $\chi$ for the second BTN as it obviously possesses less parameter complexity. However, the number of layers in the second BTN is doubled, thus possesses higher contraction complexity (e.g., deeper back propagation of the gradients). 

The expressive power of a BTN should be limited by the dimensions of the root sets $d$, the dimensions of the hidden sets $\chi$, and the network structure. Let us firstly discuss the upper bound of the expressive power. If the number of the events in a root set $d$ equals to the number of values that a feature can take, the expressive power reaches the genuine upper bound. In this case, all events are represented as orthonormal basis $\prod_{\otimes m, n} V^{[m, n]}$ of a vector space, meaning all events are mutually exclusive. There are in total $d^M$ mutually exclusive events, corresponding to $d^M$ orthonormal basis of the vector space.

However, $d$ is usually much smaller than the number of values that a feature can take. For instance, one may take $d=4$ for gray-scale images using Eq. (\ref{eq-Fmap1}) or (\ref{eq-Fmap2}), in which a pixel can take 256 values. In this case, the events that should be mutually exclusive become non-exclusive. For example, one pixel cannot simultaneously be black and gray in a given image, but the events of the pixel being black or gray are not exclusive for $d<256$. Mathematically, the vectors $V$ [Eq. (\ref{eq-Fmap1}) or (\ref{eq-Fmap2})] for the pixel being black or gray are not orthogonal to each other. In this case, the upper bound of the expressive power of the BTN is determined by $d$. 

The dimension of the hidden indexes $\chi$ determines how well the upper bound is approached. To see this, let us consider the Bayesian tensor $\mathcal{T}$ that corresponds to the whole BTN (see the property (d$'$) of the BTN). The total dimension of its in-going indexes equals to $d^M$. Such a tensor can be considered as $N_c$ normalized vectors defined in the vector space whose dimension is $d^M$. 

If one is able to properly process the full $\mathcal{T}$, in principle the upper bound can be reached. However, $\mathcal{T}$ is exponentially large. The BTN then can be regarded as an efficient approximation of the full $\mathcal{T}$ by writing it in a specific TN form. In other words, the BTN can be consider as a tensor cluster decomposed from $\mathcal{T}$ \cite{TSLS+08bayest, STS09bayest, XCH+10bayest, ZZC15bayest}, if such a decomposition exists. The dimension of the hidden indexes $\chi$ controls how accurately the BTN can approximate the $\mathcal{T}$ at the upper bound. The same argument about the expressive power on the TN machine learning has been given in Ref. \cite{LRWP+17MLTN}.

As discussed above, any connected network structure can approach the upper bound with sufficiently large $\chi$. This might be the reason that the expressive power of BTN does not severely depend on the network structure, unlike the NN's. For a BTN, the network structure should affect how fast the upper bound can be approached by increasing $\chi$. For simplicity, let us consider the loop-free structures. As we know from the TN, the fluctuations between two sets of events (formally written as $\langle xx' \rangle - \langle x \rangle\langle x' \rangle$ with $\langle x \rangle$ the average of $x$; same to the covariance) decays exponentially with the length of the path connecting these two indexes. This is a fundamental property of a loop-free TN. With a good structure, the sets with large covariance should be connected by a shorter path. Then it would require a smaller $\chi$ to capture the fluctuations among the root and leaf events.

\begin{figure}[tbp]
	\includegraphics[width=0.95\linewidth]{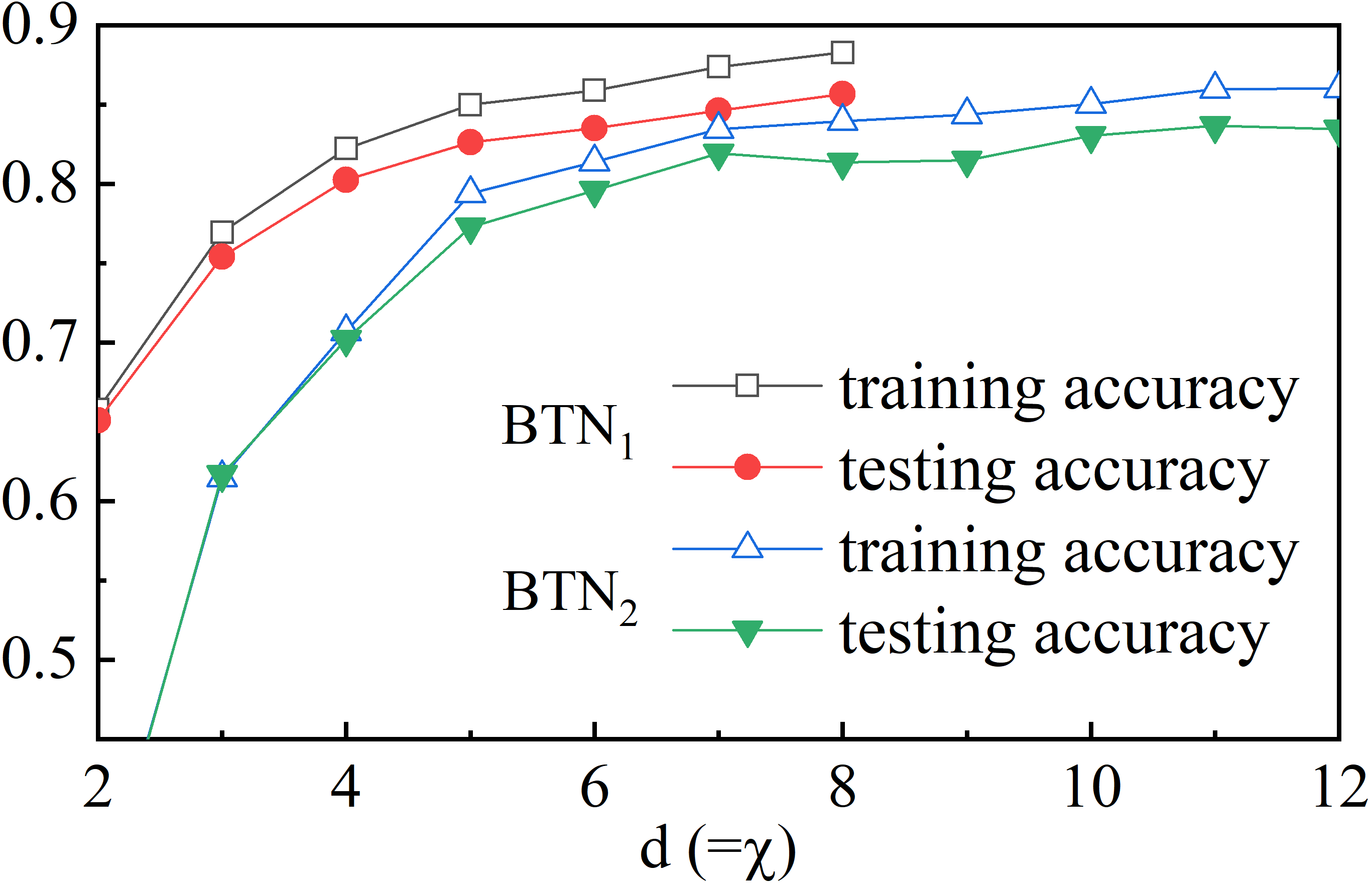}
	\caption{\label{Fig-dchi} (Color online) The training and testing accuracies against the root bond dimension $d$, while taking the hidden bond dimension $\chi =d$. About 1000 epochs are taken in each simulation.}
\end{figure}

Based on the above arguments, a conjecture about the bottleneck of the expressive power is proposed as the following: the bottleneck of a given network structure (namely structural bottleneck) is reached when the performance of the model increases exponentially slowly with the parameter complexity (see property (h) of BTN). Fig. \ref{Fig-dchi} shows the training and testing accuracies against $d$ (taking $\chi = d$). Both accuracies have not converged for the largest $d$ that our at-hand hardware can process. It means that the structural bottlenecks of such simple tree networks have not been reached yet. 

\begin{figure}[tbp]
	\includegraphics[width=1\linewidth]{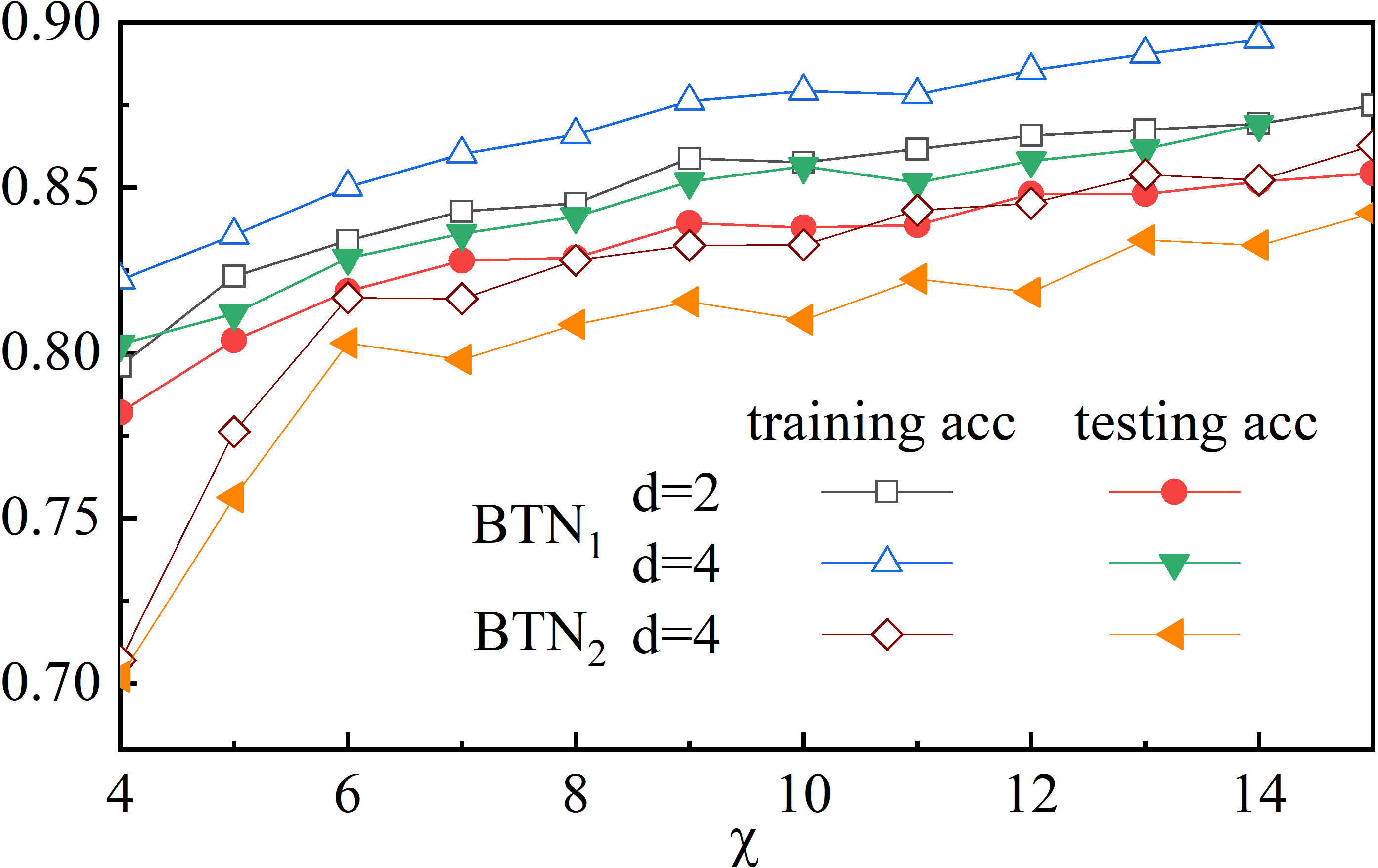}
	\caption{\label{Fig-chi} (Color online) Training and testing accuracies versus the hidden dimension $\chi$. The root dimension is taken as $d=2$ and $4$.}
\end{figure}

Fig. \ref{Fig-chi} shows the accuracies versus the hidden dimension $\chi$ with fixed root dimension ($d=2$ and 4). The accuracies increase with $\chi$, meaning the expressive power of the BTN is approaching the upper bound of the given $d$. One can see that even for $d=2$, it is not easy to reach the upper bound, as the dimension of the vector space scales as $d^M$. 

In general, BTN$_1$ outperforms BTN$_2$ while taking the same $d$ and $\chi$. There are mainly two reasons. One is that the parameter complexity of BTN$_1$ is higher than BTN$_2$ when $d$ and $\chi$ are the same. The other is that the lengths of the paths connecting two root indexes in BTN$_1$ are mostly doubled in BTN$_2$, since the depth of the tree is doubled. As the fluctuations decay exponentially with the lengths, it would require larger $\chi$ for BTN$_2$ to capture the fluctuations.

As the dimension of the vector (or sample) space is exponentially large, BBN cannot be implemented on such problems efficiently. One choice is a simplified Bayesian model called Naive Bayesian classifiers by assuming the independence of the events. BTN significantly surpasses Naive Bayesian classifiers with the testing accuracy around $70 \%$ \cite{IWAS07BayesFMNIST}. This implies that the dependences among the events are important and well-considered by BTN. In addition, it is widely accepted that probabilistic models (including BBN) are more suitable for inferences than classifications. Thus it is no surprise that BTN does not beat the state-of-the-art NN on image recognition. Note for BTN even for image recognition, there are many ways to further improve the accuracy by, e.g., pushing to larger numbers of parameters with more computational resources and/or designing more complicated and deeper network structures.

\begin{figure}[tbp]
	\includegraphics[width=1\linewidth]{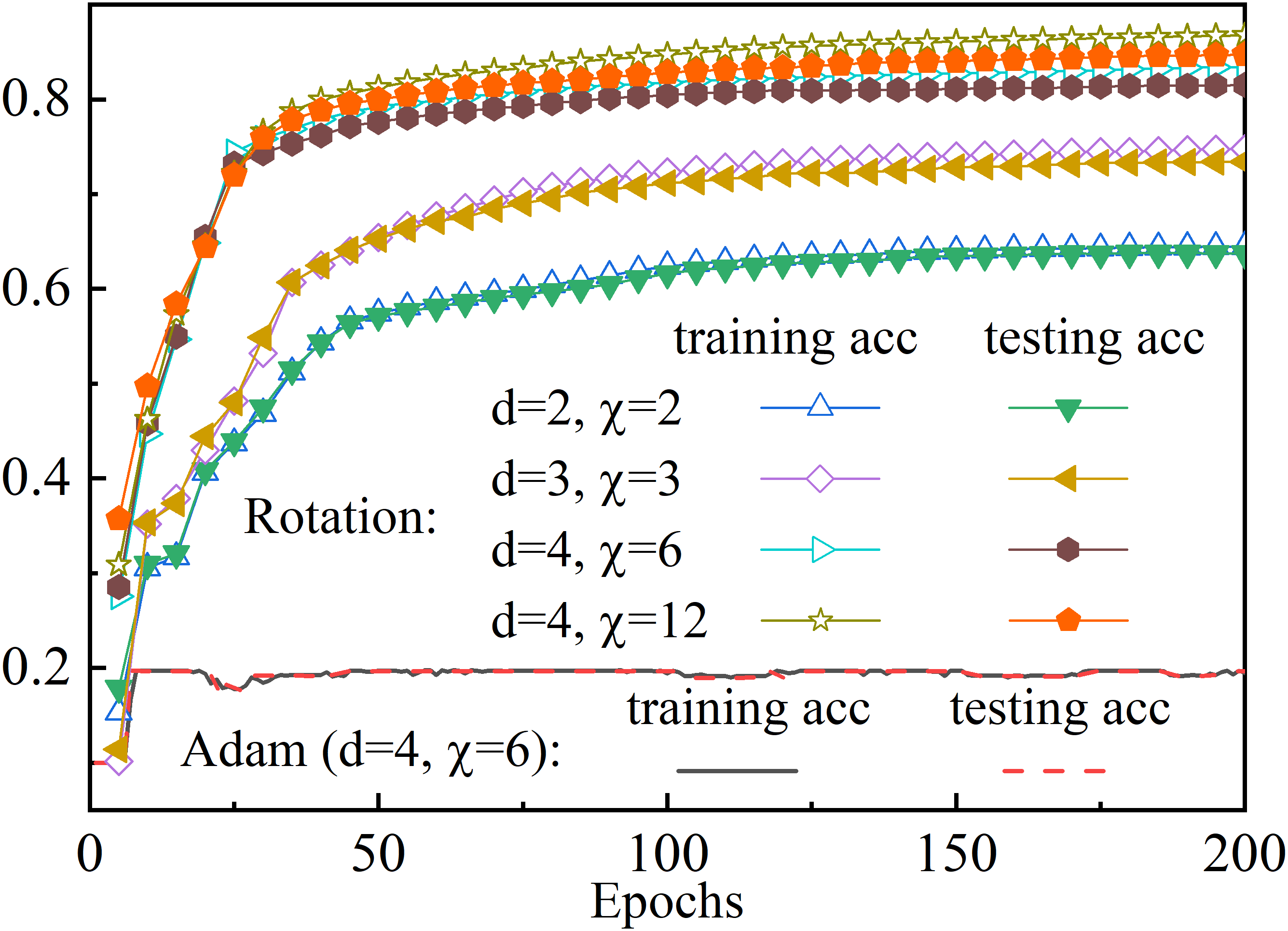}
	\caption{\label{Fig-Adam} (Color online) The training and testing accuracies for different epochs with BTN$_1$. For the rotation optimization, a fair convergence is reached after about 100 epochs. For Adam, the parameters seem to be trapped to a bad local minimum. Both learning rates are taken as $\eta = \tan \theta = 10^{-3}$.}
\end{figure}

``Area law'' means a quantity scales only with the boundary, not the bulk. The area law of entropy or entanglement appears in many systems, such as black holes \cite{B73EntAreaLaw} and quantum states \cite{ECP10AreaLawRev}. For quantum many-body simulations, a TN, including the tree TN, naturally satisfies the area law of entanglement entropy. The validity of the simple tree BTN implies an area law of covariance in the image recognition problems. If one separates the image into two parts by drawing a boundary, the main contributions of the covariance between these two parts should be from the short-range covariances of the data near the boundary. The BTN accurately captures these covariances and gives accurate results of classification. On the other hand, the performance of BTN can be improved by considering the covariances in larger ranges. Possible solutions include modifying the network structure, and pre-processing the features to localize the covariances.




Finally, Fig. \ref{Fig-Adam} demonstrates the convergences of the accuracies by the rotation optimization and Adam. For Adam, we use the standard procedure to update the tensors, and then normalize the tensors by hand to satisfy Eq. (\ref{eq-Tcond1}).  Adam seems to suffer the gradient vanishing problem and fails to give reasonable results. Note that I am not stating that Adam cannot work to optimize BTN. It is possible to obtain a good performance by carefully designing the optimization process based on Adam (e.g., by adding Lagrangian terms). From the data at hand, the rotation optimization method is more robust and efficient.

\section{Summary}

This work aims at proposing Bayesian tensor network (BTN) for probabilistic machine learning. For capturing the conditional probabilities among multiple sets of events, BTN exhibits high efficiency and accuracy with polynomial complexity when the sample space is exponentially large. An optimization method is proposed to efficiently update BTN, which avoids the potential gradient vanishing problem. Our results imply an area law of the covariance in the image recognition problems. As an efficient model, I expect that BTN will be useful not only for machine learning but also for the physical simulations by TN. 

\section*{Acknowledgment}

SJR is grateful to Ding Liu, Zheng-Zhi Sun, and Ye-Ming Meng for helpful discussions. This work was supported by Beijing Natural Science Foundation (Grant No. 1192005 and No. Z180013) and the NSFC (Grant No. 11834014), and by the Academy for Multidisciplinary Studies, Capital Normal University.

\section*{Additional information}

The code of BTN is publicly available on GitHub: \url{https://github.com/ranshiju/BayesianTN}.

%

\end{document}